\documentclass[sigconf]{acmart}

\usepackage[most]{tcolorbox}

\usepackage{makecell}

\usepackage{xspace}
\newcommand{\MultiOCR}{\texttt{MultiOCR-QA}\xspace}
\AtBeginDocument{%
  }

\copyrightyear{2025}
\acmYear{2025}
\setcopyright{cc}
\setcctype{by}
\acmConference[CIKM '25]{Proceedings of the 34th ACM International Conference on Information and Knowledge Management}{November 10--14, 2025}{Seoul, Republic of Korea}
\acmBooktitle{Proceedings of the 34th ACM International Conference on Information and Knowledge Management (CIKM '25), November 10--14, 2025, Seoul, Republic of Korea}\acmDOI{10.1145/3746252.3761295}
\acmISBN{979-8-4007-2040-6/2025/11}

\begin{document}

\title{Evaluating Robustness of LLMs in Question Answering on Multilingual Noisy OCR Data}

\author{Bhawna Piryani}
\email{bhawna.piryani@uibk.ac.at}
\affiliation{%
  \institution{University of Innsbruck}
  \city{Innsbruck}
  \country{Austria}
}
\author{Jamshid Mozafari}
\email{jamshid.mozafari@uibk.ac.at}
\affiliation{%
  \institution{University of Innsbruck}
  \city{Innsbruck}
  \country{Austria}
}
\author{Abdelrahman Abdallah}
\email{abdelrehman.abdallah@uibk.ac.at}
\affiliation{%
  \institution{University of Innsbruck}
  \city{Innsbruck}
  \country{Austria}
}
\author{Antoine Doucet}
\email{antoine.doucet@univ-lr.fr}
\affiliation{%
  \institution{University of La Rochelle}
  \city{La Rochelle}
  \country{France}
}
\affiliation{%
  \institution{University of Ljubljana}
  \city{Ljubljana}
  \country{Slovenia}
}
\author{Adam Jatowt}
\email{adam.jatowt@uibk.ac.at}
\affiliation{%
  \institution{University of Innsbruck}
  \city{Innsbruck}
  \country{Austria}
}


\begin{abstract}
Optical Character Recognition (OCR) plays a crucial role in digitizing historical and multilingual documents, yet OCR errors - imperfect extraction of text, including character insertion, deletion, and substitution can significantly impact downstream tasks like question-answering (QA). In this work, we conduct a comprehensive analysis of how OCR-induced noise affects the performance of Multilingual QA Systems. To support this analysis, we introduce a multilingual QA dataset \MultiOCR, comprising 50K question-answer pairs across three languages, English, French, and German. The dataset is curated from OCR-ed historical documents, which include different levels and types of OCR noise. We then evaluate how different state-of-the-art Large Language Models (LLMs) perform under different error conditions, focusing on three major OCR error types. Our findings show that QA systems are highly prone to OCR-induced errors and perform poorly on noisy OCR text. By comparing model performance on clean versus noisy texts, we provide insights into the limitations of current approaches and emphasize the need for more noise-resilient QA systems in historical digitization contexts.
\end{abstract}



\maketitle

\section{Introduction}
Optical Character Recognition (OCR) technology has played a crucial role in digitizing and providing access to historical texts. Over the past decade, significant advancements in OCR have improved text recognition accuracy, leading to the development of large-scale digital libraries of historical texts \cite{terras20111}. These libraries serve as valuable resources for researchers, professionals and the general public, enabling access to old manuscripts, newspapers, and other archival materials. Historical documents hold a wealth of knowledge, offering insights into past events, cultures, and people. Many professionals such as historians, journalists, or sociologists rely on these heritage collections for various research and analysis tasks.

Historical corpora have already been used as an underlying text for numerous natural language processing (NLP) tasks, including Named Entity Recognition (NER), topic modeling, text classification, neural ranking, and understanding semantic changes in language over time \cite{ehrmann2020overview, ehrmann2023named, yang-etal-2011-topic, Liebeskind2020DeepLF,subramani2020survey}. While these tasks offer valuable insights, it is expected that automatic question answering (QA) systems become an important interface to the vast historical collections. QA systems enable direct and intuitive access to information, offering a powerful means of analyzing, searching and understanding historical texts. By retrieving precise answers to user queries, QA can significantly enhance the accessibility, interpretability, and utility of historical corpora, making them more easily accessible and more useful for scholars, researchers, and the general public alike.

\begin{figure*}
    \centering
    \includegraphics[width=0.8
    \textwidth]{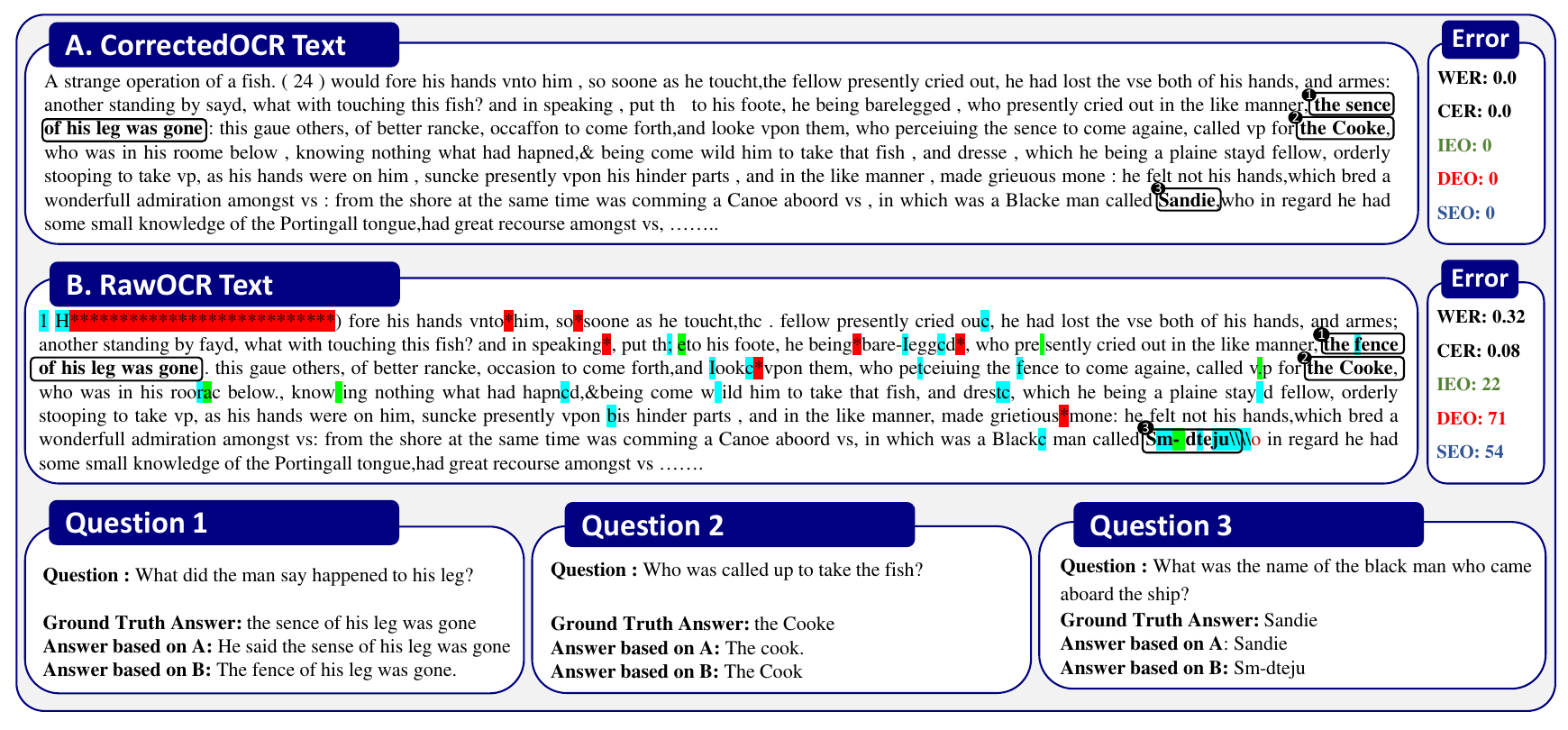}
    \caption{An example of CorrectedOCR and RawOCR text from the \MultiOCR dataset for the English language, highlighting different types of errors along with questions corresponding to this text. WER and CER denote Word Error Rate and Character Error Rate, respectively, indicating the level of errors in the text. The green highlights represent insertion errors, where IEO denotes Insertion Edit Operations - the number of insertions needed to transform RawOCR into CorrectedOCR. Similarly, red and blue highlights indicate deletion and substitution errors, with DEO and SEO representing Deletion Edit Operations and Substitution Edit Operations, respectively. The black boxes with numbers in the CorrectedOCR and RawOCR text correspond to the answers for each question in the paragraph.}
    \label{fig:example}
\end{figure*}

Despite the advancements in OCR technology, significant challenges such as character misrecognition and structural inconsistencies persist. OCR-generated text, referred as \textbf{RawOCR text} in this paper, often contains errors due to the degraded conditions and non-standard character of historical documents. Factors like faded ink and paper, irregular fonts, physical damage, and printing inconsistencies cause recognition errors that negatively impact downstream NLP tasks such as information retrieval, machine translation, and QA systems. Since QA models heavily depend on the quality of the input text, errors in RawOCR text can significantly affect the accuracy and reliability of generated answers.

For instance, in German, a passage with OCR error "\textit{Der Bericht der Lagsatzungsgesandtschaft wird verlesen undvon Hrn. Bürgermeistet Mousson als erstem Gesandten desStandes Zürich mit einigen Bemerkungen begleitet}.\footnote{English Translation: The report of the legislative mission is read out and accompanied by a few comments from Mr. Mayor Mousson, the first envoy of the Zurich state.}" contains multiple insertion, deletion, and substitution errors, such as "\textit{Lagsatzungsgesandtschaft}" (should be "Tagsatzungsgesandtschaft") and \textit{"Bürgermeistet"} (should be "Bürgermeister"). When a QA model encounters such a noisy OCR text, it may generate an incorrect answer. For example, given the question: "\textit{Wer hat den Bericht der Tagsatzungsgesandtschaft verlesen?}" (Who read the report of the parliamentary delegation?) the model incorrectly responds "\textit{Der Bericht der Tagsatzungsgesandtschaft wurde von Hrn. Bürgermeistet Mousson verlesen}." (The report of the parliamentary delegation was read out by Mr. Mayor Mousson.) which retains the OCR error and potentially misleads the QA system. This example highlights how even minor OCR errors can significantly affect QA accuracy, resulting in misleading or incorrect answers. 

Although extensive research has focused on improving OCR accuracy and post-processing correction techniques, the specific impact of OCR noise on QA remains largely unexplored. Previous studies have examined challenges related to OCR in information retrieval (IR) \cite{OCR-IR}, historical text processing \cite{piotrowski2012natural, hill2019quantifying}, and entity recognition \cite{grover-etal-2008-named}. However, a systematic investigation of how OCR errors affect QA model performance is still missing.
Furthermore, our study contributes also to the body of work on robustness of LLMs against noisy inputs such as noisy prompts  \cite{ wang2023large}, however we do it in a quite novel yet realistic setting.

In this paper, we address this gap in understanding how LLMs perform in QA tasks when dealing with noisy OCR-generated text. In particular, we first introduce \MultiOCR \footnote{The dataset is available at \url{https://github.com/DataScienceUIBK/MultiOCR-QA}},
a new multilingual QA dataset 
covering English, French, and German. This dataset includes both \textbf{RawOCR text} (OCR-generated text with errors) and \textbf{CorrectedOCR text} (ground truth), allowing for a direct comparison of QA performance under different text quality conditions. To generate contextually relevant question-answer pairs from historical text excerpts, we leverage instruction-fine-tuned LLMs. We then systematically evaluate the impact of different types of OCR errors—insertions, deletions, and substitutions—on QA model performance, offering new insights into the strengths and limitations of LLMs when processing noisy historical data. Figure~\ref{fig:example} illustrates an example from the \MultiOCR dataset with various OCR errors impacting text accuracy. It presents QA pairs and the corresponding responses for questions when using CorrectedOCR and RawOCR text as input context to Gemma 2--27b, demonstrating the effects of OCR noise on QA performance. For instance, \textit{Question 1: "What did the man say happened to his leg?"} produces different answers depending on whether clean or noisy text was input. When using the CorrectedOCR context, the model produces the response: \textit{"He said his sense of the leg was gone."} In contrast, the response based on the RawOCR context is: \textit{"The fence of his leg is gone,"} demonstrating how OCR-induced noise can significantly alter the meaning of the answer and degrade the QA accuracy.

In summary, we make the following contributions in this work:
\begin{itemize}
    \item We introduce \MultiOCR, a new multilingual QA dataset from historical texts in English, French, and German, featuring both raw and corrected OCR text, that allows direct comparison under varying text quality conditions.
    \item Using \MultiOCR, we conduct a comprehensive evaluation of LLM robustness against noisy OCR text analyzing and quantifying their impact on QA performance.
    \item We categorize different types and degrees of OCR errors for the purpose of evaluating their individual impact on QA performance. This allows providing comprehensive insights into how LLMs handle OCR-related challenges for different types and severity of errors in each of the studied languages. 
\end{itemize}

\begin{figure*}
    \centering
    \includegraphics[width=0.7\textwidth]{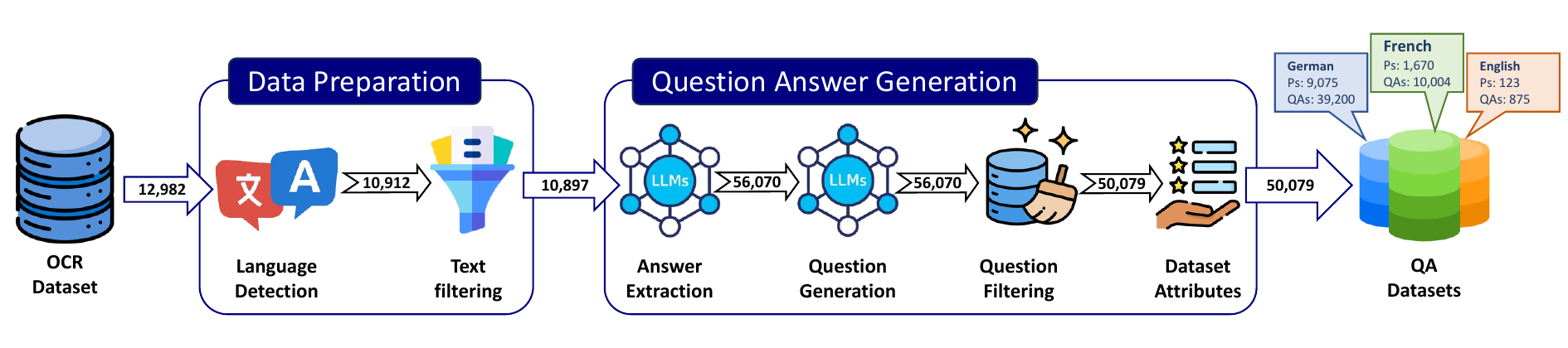}
    \caption{The Pipeline for \MultiOCR generation: Arrows represent the output quantity based on the number of documents and callouts illustrate the statistics  for English, French and German Language. Ps and QAs denote the number of paragraphs and question-answer pairs, respectively.} 
    \label{fig:QA_generation_pipeline}
\end{figure*}

\section{Related Work}


Several research works have been carried to study the limitations of OCR for historical documents and its impact on information retrieval (IR). \citet{croft1994evaluation} and \citet{traub2015impact} examined how OCR errors reduce retrieval effectiveness. \citet{chiron2017impact} found that 7\% of the relevant documents were missed due to OCR misrecognition, demonstrating the risk of failure in matching noisy texts to user queries. While these studies highlight OCR challenges, they focus primarily on document retrieval and not on question answering (QA), which requires a more fine-grained understanding of text.

Beyond IR, OCR errors have been studied in multiple tasks, including named entity recognition (NER) \cite{hamdi2020assessing, hamdi2023depth}, entity linking \cite{linhares2019impact}, text classification \cite{zu2004impact, vitman2022evaluating}, topic modeling \cite{mutuvi2018evaluating, zosa2021evaluating}, document summarization \cite{jing2003summarizing}, machine translation \cite{farooq12005effect, khayrallah-koehn-2018-impact}, and document ranking \cite{giamphy2023quantitative}. OCR noise has been shown to significantly degrade performance across these tasks. For instance, \citet{vanStrien2020AssessingTI} demonstrated that low-quality documents negatively impact multiple tasks, including dependency parsing and sentence segmentation.
\citet{hamadi20220cr} found that 80.75\% of the named entities were misrecognized due to OCR errors, causing substantial drops in accuracy. Similarly, \citet{hamdi2023depth} reported that the F1-score for NER drops from 90\% to 50\% when the character error rate increases from 2\% to 30\%. 
In topic modeling, \citet{mutuvi2018evaluating} showed that OCR noise distorts the identification of key topics. For document retrieval, \citet{OCR-IR} analyzed performance degradation at different OCR error rates, noting that retrieval effectiveness begins to decline at a word error rate of 5\% and worsens as the error rate increases. \citet{giamphy2023quantitative} further examined the impact of different types of OCR noise on document ranking and advocated for developing more robust ranking methodologies.

Despite these insights into OCR's effects on IR and NLP tasks, research on its impact on question answering remains limited. In the context of historical document collections, the only existing QA dataset, ChroniclingAmericaQA \cite{chroniclingamericaqa}, focuses primarily on creating a QA dataset from historical newspapers rather than systematically analyzing how different types of OCR errors affect QA performance. While studies on document retrieval and IR highlight OCR-related challenges, a comprehensive investigation into QA performance under different types and severity levels of OCR errors is still missing. Our work fills this gap by introducing a multilingual QA dataset (\MultiOCR) and providing a detailed evaluation of large language models (LLMs) on the RawOCR text of \MultiOCR.

\section{Methodology}
To systematically investigate the impact of OCR errors on QA systems, we constructed \MultiOCR, a new multilingual QA dataset derived from historical texts processed with OCR. This section details the two main stages of the dataset creation pipeline: Data Collection and Question-Answer Generation. Figure \ref{fig:QA_generation_pipeline} provides an overview of this process.

\subsection{Data Collection}

We first describe the process of collecting documents to generate question-answer pairs for our study. Although several historical text datasets exist, such as the IMPACT Project\footnote{\url{https://www.digitisation.eu/impact-dataset/}} and the Europeana Newspaper Project\footnote{\url{http://www.europeana-newspapers.eu/}}, which include digitized images of historical documents alongside their OCR-processed text, these lack systematically aligned ground truth corrections for a robust analysis of OCR noise effects. The ICDAR 2019 POST-OCR Text Correction dataset\footnote{\url{https://sites.google.com/view/icdar2019-postcorrectionocr}}~\cite{icdar-2019} is an exception here as it provides both RawOCR text along with its aligned ground truth (called CorrectedOCR), making it especially suitable for our research objectives. Furthermore, several English, French, and German document sources included in the IMPACT and Europeana projects are already incorporated into the ICDAR 2019 dataset, offering a diverse collection of cleaned resources for historical OCR-based studies.

ICDAR 2019 dataset includes over 22 million OCR-processed characters and their corresponding, aligned ground truth across several European languages. We focus on English, French, and German languages due to several reasons. First, French and German are among the most well-represented languages in the ICDAR 2019 dataset regarding their document numbers. Second, all three languages are high-resource and well-supported by the current large language models (LLMs), which increases the reliability of QA generation and evaluation \cite{li2025language}. Since noisy OCR text is expected to pose already a significant challenge for LLMs, incorporating low-resource languages, where LLM performance tends to be generally weaker, might introduce additional complexity. Third, each of these languages has an existing, publicly available QA dataset like SQuAD \cite{rajpurkar-etal-2016-squad} for English, FQuAD \cite{dhoffschmidt-etal-2020-fquad} for French, and GermanQuAD \cite{moller-etal-2021-germanquad} for German, which enable effective instruction fine-tuning of LLMs for high-quality QA pair generation. Therefore, we limited our study to these three languages, intending to expand to low-resource and typologically diverse languages in the future work.

\textbf{\textit{Language Specific Data Collection}:} 
The texts in ICDAR 2019 dataset originally came from 
various historical document repositories. 
\begin{itemize}
    \item \texttt{English:} The documents for the English language in the ICDAR 2019 dataset are sourced from IMPACT - British Library, comprising a total of 150 files. 
    \item \texttt{French:} For the French language, the ICDAR 2019 dataset provides a collection of 2,800 files obtained from three sources: the HIMANIS\footnote{\url{https://www.himanis.org}} Project, IMPACT - National Library of France, and the RECEIPT\footnote{\url{http://findit.univ-lr.fr/}} dataset. 
    \item \texttt{German:} The German-language dataset includes the OCR-processed text from multiple sources, such as, front pages of the Swiss newspaper NZZ\footnote{\url{https://zenodo.org/records/3333627}}, IMPACT - German National Library, GT4Hist-dta19 dataset, GT4Hist - EarlyModernLatin, GT4Hist - Kallimachos, GT4Hist - RefCorpus-ENHG-Incunabula, and GT4Hist - RIDGES-Fraktur\footnote{\url{https://zenodo.org/records/1344132}} \cite{springmann2018gt4hist}. The German dataset in ICDAR 2019 originally contained 10,032 files. 
\end{itemize}

\textbf{\textit{Language Verification and Filtering}:} Prior to QA pair generation, we preprocessed the ICDAR dataset to ensure that each document contains text in the correct target language. We applied langdetect\footnote{\url{https://github.com/Mimino666/langdetect}} library to detect the language of documents. 
The analysis revealed that some documents labeled as English, French and German were actually in other languages, particularly Latin. To maintain dataset integrity, we removed non-target language documents, resulting in the following reductions: We removed non-English documents, reducing the number of documents in the English dataset to 141. Similarly, we discarded non-French documents, reducing the dataset to 1,713 French-language files, and eliminating 1,086 Latin-language files. Finally, for German, we removed Latin or other non-German files and retained a total of 9,075 German-language files.

Furthermore, the ICDAR dataset, originally intended for post-OCR correction, contained special alignment symbols (e.g., \texttt{@}, \texttt{\#}) to map the RawOCR text to its ground-truth counterpart. We removed these symbols from the ground-truth text before generating questions. We also excluded files where the ground-truth text was missing, resulting in the removal of 16 files for English, 3 files for French, and none for German. This preprocessing step ensured that all QA pairs were generated from text that had both CorrectedOCR text and RawOCR text.

\subsection{Question-Answer Generation}

To construct the multilingual QA dataset, we opted for automatic QA pair generation, as manual dataset creation would require substantial human resources. To achieve this, we instruction fine-tuned a pretrained LLM for each target language to generate QA pairs from CorrectedOCR text.

While LLMs are pretrained on diverse NLP tasks, they typically generate a variety of question types, including non-factoid and open-ended questions. Since our goal is to develop a factoid QA dataset, we fine-tuned the models using language-specific QA datasets to ensure the generation of structured, precise, and factual question-answer pairs.

Instruction fine-tuning enhances both the capabilities and controllability of LLMs \cite{zhang2023instruction}. Fine-tuning instruction-based datasets across multiple languages allows the model to generalize across different question-answering styles, ensuring that the generated questions remain relevant even when dealing with language-specific variations. We opted to finetune LLaMa-3.1-70B instruct model \cite{dubey2024llama} separately for each language using widely adopted QA datasets.

 For fine-tuning the model in English, we used the SQuAD v1 dataset \cite{rajpurkar-etal-2016-squad}. We randomly selected 2,067 paragraphs and 10,570 questions from the development set and 3,000 paragraphs and 13,894 questions from the test set. 
 
 Similarly, for the French language, we fine-tuned the model on the FQuAD dataset \cite{dhoffschmidt-etal-2020-fquad}, utilizing both the validation and training sets, comprising 5,689 paragraphs and 23,919 questions. For the German language, we fine-tuned the model on the GermanQuAD dataset \cite{moller-etal-2021-germanquad} using the training and validation sets, which consisted of 3,014 paragraphs and 13,722 questions. This fine-tuning step ensured that the model accurately generated factoid-style QA pairs, reducing instances of open-ended or conversational questions.

After instruction fine-tuning, we used the fine-tuned model for each language to generate questions for the preprocessed dataset prepared in the initial step. To 
generate high-quality QA pairs, we employed a \textbf{two-step prompt-based approach:} Answer  Extraction and Question Generation for the Extracted Answers. 

\textbf{\textit{Answer Extraction:}} The model was first prompted to extract multiple candidate answer spans from a given passage. These spans included entities, numbers, dates, locations, and key phrases that could serve as factual answers. The following prompt was used for the extraction of the candidate answer.

\begin{tcolorbox}[size=small,colback=blue!2!white,colframe=blue!50!black, title=English Answer Extraction Prompt ]
\begin{quote}
\footnotesize
\emph{\textit{\textbf{System Prompt:}} You are an expert at extracting key information from text. Your goal is to identify spans of text that are likely to serve as answers to potential questions based on the input passage. Focus on meaningful, distinct, and diverse snippets such as entities, nouns, verbs, adjectives, numbers, dates, and phrases. Avoid redundancy and ensure the answers are diverse, representing key information in the passage.}

\emph{\textbf{\textit{User Prompt}}: Given the passage below, extract several candidate spans that are likely to be answers to potential questions. Write only the extracted answers, separated by a semicolon (;). Passage : \{context\}}
\end{quote}
\end{tcolorbox}

After generating the candidate answers, we checked for duplicated answer spans. If the answers were duplicate, we removed them and retained only unique answers. This prompt was applied uniformly across all three languages, with translations adapted to each language.

\textbf{\textit{Question Generation:}} Following answer extraction, the extracted answer spans were re-fed into the model, and it was prompted to generate questions that align with each answer while maintaining contextual relevance. The generated questions were structured to be standalone, well-formed, and factually grounded in the passage. The following Prompt was used for Question Generation from the Extracted Answers.

\begin{tcolorbox}[size=small,colback=blue!2!white,colframe=blue!50!black, title=English Question Generation Prompt ]
\begin{quote}
\footnotesize
\emph{\textit{\textbf{System Prompt:}} You are an expert at generating standalone questions based on the provided passage. Your goal is to create a clear, relevant, and self-contained question that aligns with the information in the passage. The question should not explicitly reference the passage or require additional information to be understandable. Ensure the question is concise, well-structured, and meaningful.}

\emph{\textit{\textbf{User Prompt:}} Based on the passage below, please generate a question that is relevant to the information provided. The question should be standalone, clear, and understandable without referencing the passage directly. The answer to the question should be [{answer}]. Passage : \{context\}}
\end{quote}
\end{tcolorbox}

Using this approach, we generated 941 questions for English, 10,522 questions for French, and 44,607 questions for German.

\textit{\textbf{Dataset Filtering:}} After generating the dataset for each language, we applied additional filtering steps to ensure quality and consistency. Specifically, we removed questions that did not end with a question mark, duplicate questions, and questions with excessively long answers. Since LLM-generated datasets may contain hallucinated long answers, we applied this additional filtering by removing excessively long answers. 
As a result, we removed 66 questions for English Language, 530 questions for French Language, and 7,210 questions for German Language. This final filtering step ensured that the \MultiOCR dataset consisted of concise, well-structured, and factually accurate question-answer pairs.

\section{Dataset Analysis}

After applying all the filtering steps, we obtained the final dataset, comprising 50,079 question-answer pairs. The dataset statistics, including average paragraph length, question length, and answer length, are presented in Table \ref{tab:dataset_statistics}.

\begin{figure*}
    \centering
    \includegraphics[width=0.65\textwidth]{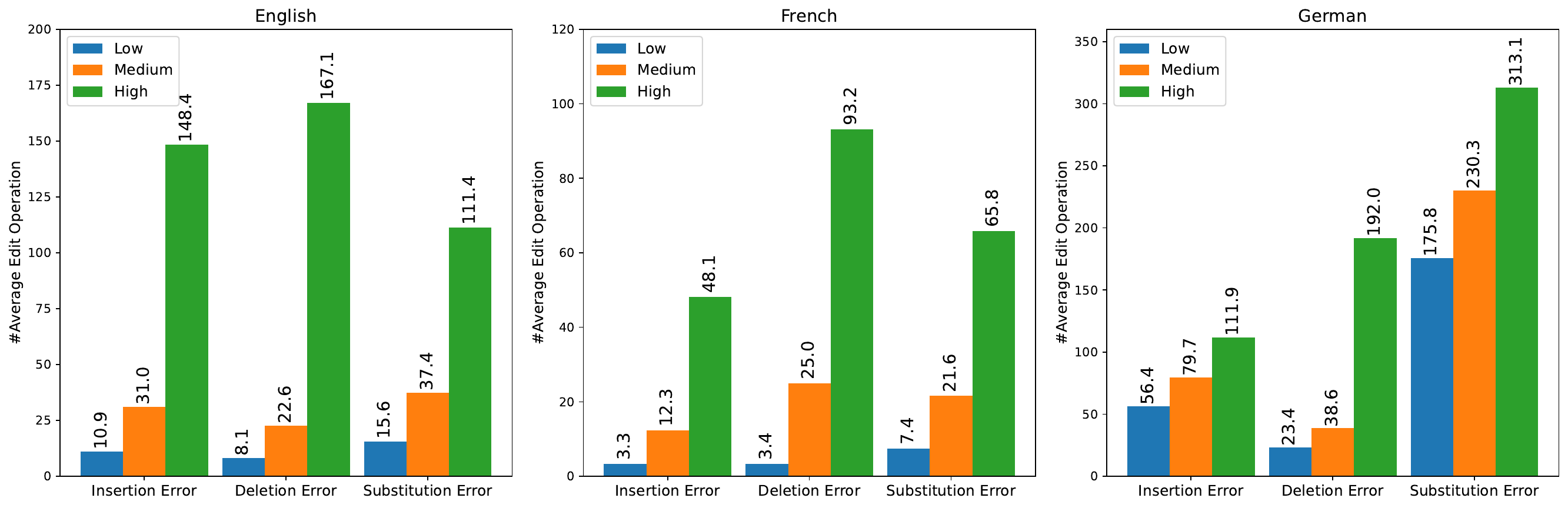}
    \caption{Statistics of insertion, deletion, and substitution errors for each language, categorized into low, medium, and high noise levels.}
    \label{fig:error_type_edit_ops}
\end{figure*}

\subsection{Quantifying and Filtering Noise}
\label{error_quantification}

To assess the impact of OCR noise on QA quality, we quantified the noise level in the RawOCR text using two standard metrics: Character Error Rate (CER) and Word Error Rate (WER). CER measures the proportion of character-level errors in the RawOCR text compared to the ground truth text. It is computed as the number of insertions, deletions, and substitutions (including spaces) required to transform the RawOCR text into its correct form. WER quantifies word-level discrepancies, representing the proportion of words that require modifications (insertions, deletions, or substitutions) to match ground truth. Both CER and WER were computed using the Levenshtein distance \cite{miller-2009}, which determines the minimum number of edits needed to correct the OCR-generated text. A high CER but low WER suggests that errors are concentrated within a few words (e.g., spelling variations), whereas a high WER indicates distortions across multiple words, significantly affecting readability.

\begin{table}[t]
\centering
\caption{Basic statistics of the \MultiOCR dataset, including question-answer (QA) pair count, paragraph count, and average text lengths across languages.}
\label{tab:dataset_statistics}
\resizebox{0.75\columnwidth}{!}{%
\begin{tabular}{cccc}
\toprule
                                            & \textbf{English}   & \textbf{French}    & \textbf{German}   \\
\toprule
\#QA pairs                                       & 875  & 10,004   & 39,200  \\
\#Paragraphs                                     & 123   & 1,670    &9,075 \\
Avg. CorrectedOCR paragraph length (words)       & 271.73  & 297.53   & 217.33 \\
Avg. RawOCR paragraph length (words)             & 263.46  & 335.73   &193.23 \\
Avg. question length (words)                     & 8.60   & 8.73     & 8.08 \\
Avg. answer length (words)                       & 2.93    & 3.12     & 5.63 \\
Avg. questions per paragraph                     & 7.11    & 5.99     & 4.32   \\
\bottomrule
\end{tabular}%
}
\end{table}

\textbf{\textit{Outlier Detection:}} To ensure a reliable analysis, we applied the Interquartile Range (IQR) method to detect and remove outliers in CER values. Outliers were defined as CER values below Q1 - 1.5 * IQR or above Q3 + 1.5 * IQR, where Q1 and Q3 represent the 33rd and 66th percentiles of the CER distribution, respectively, and IQR is the difference between Q3 and Q1. This filtering resulted in the removal of 25 English, 351 French, and 2,423 German paragraphs.

Following outlier removal, we categorized the remaining paragraphs into three noise levels based on CER percentiles for each language. Documents with CER below the 33rd percentile were classified as "low noise," those between the 33rd and 66th percentiles as "medium noise", and those above the 66th percentile as "high noise." The specific CER thresholds for each category and language were as follows:

\begin{itemize}
    \item \texttt{English:} Low: CER $<$ 0.0576, Medium: $0.0576 \leq$ CER $<$ 0.1238, High: CER $\geq$ 0.1238
    \item \texttt{French:} Low: CER $<$ 0.0357, Medium: $0.0357 \leq$ CER $<$ 0.0558, High: CER $\geq$ 0.0558
    \item \texttt{German:} Low: CER $<$ 0.2453, Medium: $0.2453 \leq$ CER $<$ 0.2757, High: CER $\geq$ 0.2757
\end{itemize}

\begin{table}[t]
\centering
\caption{OCR Error Statistics across Languages}
\label{tab:ocr_error_statistics}
\resizebox{0.6\columnwidth}{!}{
\begin{tabular}{lccc}
\toprule
\textbf{Metric} & \textbf{English} & \textbf{French} & \textbf{German} \\
\midrule
\multicolumn{4}{c}{\textbf{Character Error Rate (CER)}} \\
Mean   & 0.1245  & 0.0519  & 0.2816  \\
Median & 0.0729  & 0.0440  & 0.2592  \\

\midrule
\multicolumn{4}{c}{\textbf{Word Error Rate (WER)}} \\
Mean   & 0.3047  & 0.1904  & 0.8713  \\
Median & 0.2711  & 0.1760  & 0.8730  \\

\midrule
\multicolumn{4}{c}{\textbf{Edit Operations}} \\
Mean Substitutions      & 54.97  & 31.51  & 240.23  \\
Median Substitutions    & 38.00  & 21.00  & 230.50  \\
Mean Deletions        & 64.42  & 41.09  & 85.09  \\
Median Deletions        & 18.00   & 25.00  & 39.00  \\
Mean Insertions       & 66.91  & 20.90  & 82.60  \\
Median Insertions       & 27.50  & 11.00  & 80.00  \\

\bottomrule
\end{tabular}
}
\end{table}

In addition to CER-based classification, we analyzed the distribution of three specific OCR error types: insertions, deletions, and substitutions. Each error type was categorized separately using a percentile-based approach, allowing for a more detailed examination of the nature and severity of OCR distortions. To further investigate OCR noise patterns, we classified insertion, deletion, and substitution errors into low, medium, and high noise levels.
As illustrated in Figure \ref{fig:error_type_edit_ops}, the distribution of these error types varies significantly across languages, reflecting differences in OCR quality and text processing challenges in English, French, and German. Additionally, Table \ref{tab:ocr_error_statistics} presents the statistical characteristics of the distribution of OCR error metrics across languages, including CER, WER, and edit operations, providing further insights into the OCR noise characteristics.

\section{Experiments and Results}

In this section, we conduct a comprehensive analysis of \MultiOCR from several perspectives. First, we evaluate the performance of \MultiOCR across various LLMs, comparing different model families and sizes to assess their effectiveness in handling OCR-generated text. Second, we investigate the impact of OCR errors on QA performance, focusing on three main error types: insertion errors, deletion errors, and substitution errors. 

\vspace{-2mm}

\subsection{Experimental Settings}

We conducted experiments using multiple large language models (LLMs), including Qwen2.5 7B \cite{yang2024qwen2}, LLaMa 3.1-8B \cite{dubey2024llama}, Gemma-2-27B \cite{team2024gemma}, Mixtral 8x22B \cite{jiang2024mixtral}, LLaMA 3.3-70B \cite{dubey2024llama}, and Qwen2.5 72B \cite{yang2024qwen2}. These models span different architectures and parameter sizes, allowing for a comprehensive comparison on OCR text.

Traditionally, QA systems are evaluated using Exact Match (EM). However, these metrics can be insufficient for LLMs, as models often generate verbose responses, leading to low EM scores even when the correct answer is included in the response. To address this limitation, we evaluate \MultiOCR using \textbf{BERTScore} \cite{bertscore} alongside \textbf{EM}. Additionally, we introduce another evaluation metric, \textbf{Contains}, to better assess \MultiOCR performance. Contains measures the extent to which the ground truth is present in the response generated by the model, regardless of the verbosity. 

We will apply these metrics to evaluate the QA results on RawOCR texts and then on CorrectedOCR texts used as context.

\subsection{Experimental Results}
\label{experimental_results}
Tables \ref{tab:model_performance_LLM_English}, \ref{tab:model_performance_LLM_French}, and \ref{tab:model_performance_LLM_German} present the impact of OCR errors on the performance of various LLMs in question-answering tasks for English, French, and German texts. 

\textbf{English text:}
Table \ref{tab:model_performance_LLM_English} presents the performance of LLMs on English text using both CorrectedOCR (CP) and RawOCR (RP) paragraphs. Across all models, the transition from CP to RP negatively impacts performance. The best-performing model on CP is LLaMA-3.3-70B with a BERTScore of 66.94, followed by Gemma-2 27B at 63.99. When switching to RP, LLaMA-3.3-70B still achieves the highest BERTScore 62.87, but with a 6.08\% drop, highlighting its robustness. The lowest-performing model is Mixtral 8x22B, showing the most significant impact of OCR errors. LLaMA-3.3-70B model achieves the best performance in CP for the Contains metric of 63.90, indicating its strong retrieval ability in clean text. However, it experiences a 21.3\% drop in RP, suggesting moderate sensitivity to OCR errors. The lowest performing model in RP is again Mixtral 8x22B for the Contains metric, which sees a 15.70\% decrease, indicating its greater vulnerability to noisy text. The EM metric shows the steepest decline across models, emphasizing that OCR errors severely impact the models' ability to generate precise answers.
 
\textbf{French text}:
In Table \ref{tab:model_performance_LLM_French} we summarize how LLMs perform on QA over French text.  
Models consistently perform better on CorrectedOCR, confirming and quantifying the negative influence of OCR errors on the QA accuracy. Gemma-2 27B model achieves the highest BERTScore of 76.51 on CP, demonstrating its strong ability to capture semantic similarity. Despite the 1.46\% drop, it still maintains the highest performance with a BERTScore of 75.39 on RP, indicating robustness to OCR noise. Mixtral 8x22B model shows the smallest drop of 1.05\%, but its overall score remains lower than the ones for the other models.  
For Contains, Qwen-2.5 72B achieves the highest score on CP 57.55, highlighting superior retrieval performance on clean text. However, it experiences a 19.90\% drop in RP, reinforcing its vulnerability to OCR errors. LLaMA-3.1 8B also struggles, with a 19.63\% decline, showing difficulty in retrieving spans from noisy text. In terms of EM, Gemma-2 27B outperforms all models with an EM of 17.46 on CP. Although it drops by 15.23\%, it still maintains the best EM value 14.80 in RP. 
Mixtral 8x22B model is the most affected in EM, dropping by 27.99\%, suggesting that OCR noise drastically reduces its accuracy in generating exact answers.

\begin{table}[t]
\caption{Performance of LLMs on English Language: Comparison of CorrectedOCR (CP) and RawOCR Paragraphs (RP) as Context. Red numbers indicate the percentage decrease in performance with RP. Bold values highlight the highest performance for each metric with CP, while underlined values denote the best performance for each metric with RP.}
\label{tab:model_performance_LLM_English}
\centering
\resizebox{\columnwidth}{!}{
\begin{tabular}{lclll}
\toprule
\textbf{Model} & \textbf{Parameter} & \textbf{BERTScore} & \textbf{Contains}  & \textbf{EM} \\
\midrule
Qwen-2.5 (CP) & 7B & 53.28 &60.04 &5.21 \\
Qwen-2.5 (RP) & 7B & 50.12 \textcolor{red}{ (5.93\%$\downarrow$) } & 45.95 \textcolor{red}{ (23.46\%$\downarrow$) }  & 3.67 \textcolor{red}{ (29.56\%$\downarrow$) } \\
\midrule
LLaMA-3.1 (CP) & 8B & 55.39 &63.71  &0.97 \\
LLaMA-3.1 (RP) & 8B & 52.18 \textcolor{red}{ (5.80\%$\downarrow$)}  & \underline{51.74} \textcolor{red}{ (18.78\%$\downarrow$) }  & 0.39 \textcolor{red}{ (59.79\%$\downarrow$) } \\
\midrule
Gemma-2 (CP) & 27B & 63.99  &58.09  &\textbf{16.76} \\
Gemma-2 (RP) & 27B & 58.08 \textcolor{red}{ (9.23\%$\downarrow$) } & 46.20 \textcolor{red}{ (20.48\%$\downarrow$) }  & \underline{9.36} \textcolor{red}{ (44.15\%$\downarrow$) } \\
\midrule
Mixtral (CP) & 8x22B & 45.83   &57.72 &0.77 \\
Mixtral (RP) & 8x22B & 44.72 \textcolor{red}{ (2.42\%$\downarrow$) } & 48.65\textcolor{red}{ (15.70\%$\downarrow$) }  & 0.58 \textcolor{red}{ (24.68\%$\downarrow$) } \\
\midrule
LLaMA-3.3 (CP) & 70B & \textbf{66.94}  &\textbf{ 63.90}  &11.39\\
LLaMA-3.3 (RP) & 70B & \underline{62.87}\textcolor{red}{ (6.08\%$\downarrow$) }& 50.39 \textcolor{red}{ (21.3\%$\downarrow$) } & 8.11 \textcolor{red}{ (28.80\%$\downarrow$) } \\
\midrule
Qwen-2.5 (CP) & 72B & 53.69   &63.71   &7.92\\
Qwen-2.5 (RP) & 72B &50.53 \textcolor{red}{ (5.88\%$\downarrow$) } & 50.19 \textcolor{red}{ (21.24\%$\downarrow$) }  & 4.44 \textcolor{red}{ (43.94\%$\downarrow$) } \\
\bottomrule
\end{tabular}
}
\end{table}

\begin{table}[t]
\caption{Performance of LLMs on French Language: Comparison Using CorrectedOCR (CP) and RawOCR Paragraphs (RP) as Context. Red numbers indicate the percentage decrease in performance with RP. Bold values highlight the highest performance for each metric with CP, while underlined values denote the best performance for each metric with RP.}
\label{tab:model_performance_LLM_French}
\centering
\resizebox{\columnwidth}{!}{
\begin{tabular}{lclll}
\toprule
\textbf{Model} & \textbf{Parameter} & \textbf{BERTScore} & \textbf{Contains}  & \textbf{EM} \\
\midrule

Qwen-2.5 (CP) & 7B & 73.12 & 54.82 & 11.38 \\
Qwen-2.5 (RP) & 7B & 72.03 \textcolor{red}{ (1.49\%$\downarrow$) } & 42.99 \textcolor{red}{ (21.58\%$\downarrow$) } & 9.59 \textcolor{red}{ (15.75\%$\downarrow$) } \\
\midrule
LLaMA-3.1 (CP) &8B & 70.16 & 51.35 & 0.64 \\
LLaMA-3.1 (RP) & 8B & 69.30 \textcolor{red}{ (1.23\%$\downarrow$) } & 41.27 \textcolor{red}{ (19.63\%$\downarrow$) } & 0.61 \textcolor{red}{ (5.56\%$\downarrow$) } \\
\midrule
Gemma-2 (CP) & 27B & \textbf{76.51} & 52.58 & \textbf{17.46} \\
Gemma-2 (RP) & 27B & \underline{75.39} \textcolor{red}{ (1.46\%$\downarrow$) } & 42.05 \textcolor{red}{ (20.02\%$\downarrow$) } & \underline{14.80} \textcolor{red}{ (15.23\%$\downarrow$) } \\

\midrule
Mixtral (CP) & 8x22B & 68.65 & 48.32 & 0.30 \\
Mixtral (RP) & 8x22B & 67.93 \textcolor{red}{ (1.05\%$\downarrow$) } & 38.96 \textcolor{red}{ (19.37\%$\downarrow$) } & 0.21 \textcolor{red}{ (27.99\%$\downarrow$) } \\
\midrule
LLaMA-3.3 (CP) & 70B & 72.50 & 54.00 & 4.41 \\
LLaMA-3.3 (RP) & 70B & 71.42 \textcolor{red}{ (1.49\%$\downarrow$) } & 43.22 \textcolor{red}{ (19.96\%$\downarrow$) } & 3.89 \textcolor{red}{ (11.72\%$\downarrow$) } \\
\midrule
Qwen-2.5 (CP) & 72B & 73.26 & \textbf{57.55} & 10.62 \\
Qwen-2.5 (RP) & 72B & 71.98 \textcolor{red}{ (1.75\%$\downarrow$) } & \underline{46.10} \textcolor{red}{ (19.90\%$\downarrow$) } & 7.84 \textcolor{red}{ (26.16\%$\downarrow$) } \\
\bottomrule
\end{tabular}
}
\end{table}

\begin{table}[t]
\caption{Performance of LLMs on German Language: Comparison Using CorrectedOCR (CP) and RawOCR Paragraphs (RP) as Context. Red numbers indicate the percentage decrease in performance with RP. Bold values highlight the highest performance for each metric with CP, while underlined values denote the best performance for each metric with RP.}
\label{tab:model_performance_LLM_German}
\centering
\resizebox{\columnwidth}{!}{
\begin{tabular}{lclll}
\toprule
\textbf{Model} & \textbf{Parameter} & \textbf{BERTScore} & \textbf{Contains}  & \textbf{EM} \\
\midrule
Qwen-2.5 (CP) & 7B & 62.76 & 15.88 & 0.389 \\
Qwen-2.5 (RP) & 7B & 59.52 \textcolor{red}{ (5.16\%$\downarrow$) } & 5.33 \textcolor{red}{ (66.40\%$\downarrow$) } & 0.192 \textcolor{red}{ (50.45\%$\downarrow$) } \\
\midrule
LLaMA-3.1 (CP) & 8B & 63.87 & 15.36 & 0.457 \\
LLaMA-3.1 (RP) & 8B & 58.29 \textcolor{red}{ (8.74\%$\downarrow$) } & 5.31 \textcolor{red}{ (65.37\%$\downarrow$) } & 0.135 \textcolor{red}{ (70.31\%$\downarrow$) } \\
\midrule
Gemma-2 (CP) & 27B & \textbf{67.07} & 11.56 & \textbf{2.691} \\
Gemma-2 (RP) & 27B & \underline{63.78} \textcolor{red}{ (4.91\%$\downarrow$) } & 4.47 \textcolor{red}{ (61.27\%$\downarrow$) } & \underline{0.564} \textcolor{red}{ (79.04\%$\downarrow$) } \\
\midrule
Mixtral (CP) & 8x22B & 60.87 & 11.23 & 0.078 \\
Mixtral (RP) & 8x22B & 58.14 \textcolor{red}{ (4.48\%$\downarrow$) } & 4.56 \textcolor{red}{ (59.37\%$\downarrow$) } & 0.021 \textcolor{red}{ (72.74\%$\downarrow$) } \\
\midrule

LLaMA-3.3 (CP) & 70B & 63.82 & \textbf{17.68} & 0.553 \\
LLaMA-3.3 (RP) & 70B & 58.44 \textcolor{red}{ (8.43\%$\downarrow$) } & 6.24 \textcolor{red}{ (64.72\%$\downarrow$) } & 0.203 \textcolor{red}{ (63.23\%$\downarrow$) } \\
\midrule
Qwen-2.5 (CP) & 72B & 63.25 & 16.43 & 0.699 \\
Qwen-2.5 (RP) & 72B & 60.04 \textcolor{red}{ (5.08\%$\downarrow$) } & \underline{6.50} \textcolor{red}{ (60.44\%$\downarrow$) } & 0.167 \textcolor{red}{ (76.02\%$\downarrow$) } \\
\bottomrule
\end{tabular}
}
\end{table}

\begin{figure*}
    \centering
    \includegraphics[width=0.75\textwidth]{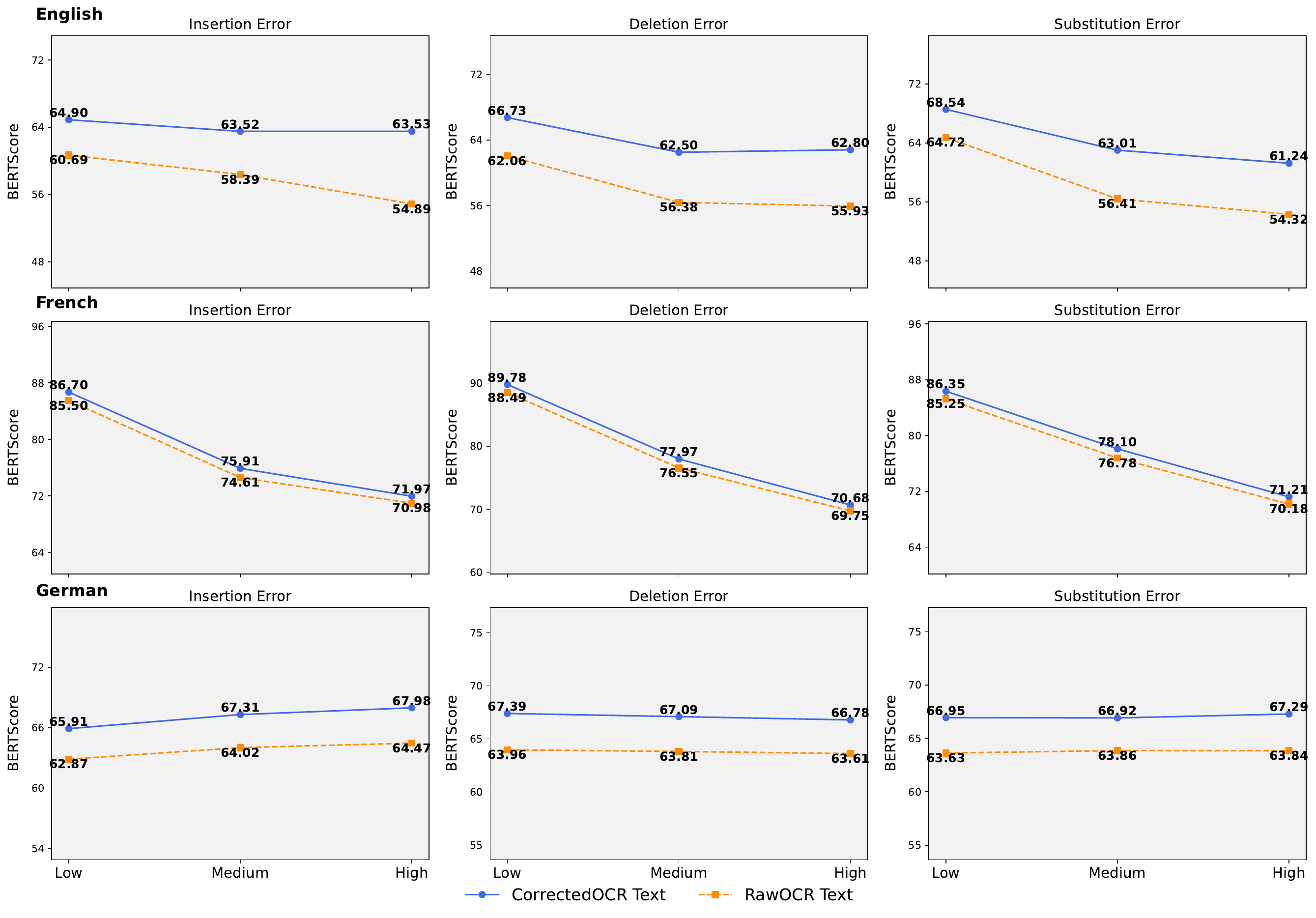}
    \caption{BERTScore of different error types on Low, medium and High categories for each Language in \MultiOCR dataset. }
    \label{fig:f1_vs_error_type}
\end{figure*}

\textbf{German text: }
Table \ref{tab:model_performance_LLM_German} focusing on German language shows the most significant performance decline among the three languages. Unlike English and French, German exhibits the largest performance drop due to its lower OCR quality, and the models struggle more with its linguistic structure. 
Among the models evaluated, Gemma-2 27B achieves the highest BERTScore of 67.07 on CP, confirming its strong ability to capture semantic similarity in clean text. It also maintains the best performance on RP (63.78); however, it still experiences a 4.91\% decrease, highlighting the adverse effects of OCR errors.
In contrast, LLaMA-3.1 8B and LLaMA-3.3 70B show the biggest BERTScore drop (8.74\% and 8.43\% respectively), suggesting that these models struggle more with OCR noise. 
For the Contains metric, LLaMA-3.3 70B achieves the highest Contains score on CP, making it the most effective for retrieving relevant information in clean text. However, all models suffer a severe drop in Contains when moving to RP, with reductions exceeding 65\%. Qwen-2.5 7B has the worst drop 66.40\%, indicating that it faces challenges to retrieve information from noisy text. In terms of EM, Gemma-2 27B achieves the highest EM score on CP 2.69, while Mixtral 8x22B has the lowest EM 0.078. EM scores drop drastically across all models, the largest decrease is 79.04\% for Gemma-2, reinforcing that word-level distortions from OCR errors make exact answer matching nearly impossible. The Mixtral model drops 72.74\% in EM, making it highly unreliable for exact answers in noisy OCR text.

\textbf{Summary of findings:} OCR errors consistently degrade the performance of the models in English, French and German texts, resulting in maximum drop of 9.23\%, 1.75\%, and 8.74\% in BERTScore, respectively. 
The most severe impact was observed in German due to the lower quality of the OCR and the complex linguistic structure. While larger models like Gemma-2 27B and LLaMA-3.3 70B demonstrate greater resilience, all models suffer substantial declines in Contains and Exact Match (EM) metrics, highlighting their weakness in retrieving and generating precise answers from noisy text. Gemma-2 27B consistently outperforms others, maintaining the highest BERTScore and EM across all languages, but still experiences notable degradation in noisy conditions. Mixtral 8x22B emerges as the most vulnerable, exhibiting the lowest performance and struggling particularly with exact answer generation.

\vspace{-2mm}
\subsection{Performance based on Different Error Types}
In this section, we conduct an in-depth analysis of the impact of different types of OCR errors: insertion, deletion, and substitution on QA systems. We use Gemma-2 (27B) for this analysis, as it 
consistently outperform the other models across English, French, and German. As detailed in Section \ref{error_quantification}, each error type has been categorized into three levels: Low, Medium, and High, where Low represents minimal presence and High indicates the most frequent occurrence of a particular type of error. We evaluated \MultiOCR's performance across these categories, as illustrated in Figure \ref{fig:f1_vs_error_type}.

Insertion errors introduce extraneous characters or words, leading to moderate performance degradation. At low and medium insertion levels, the effect on BERTScore remains relatively minor, suggesting that small insertions do not always disrupt semantic meaning. However, at high insertion levels, performance drops sharply, indicating that excessive insertions impair both readability and semantic coherence.

Deletion errors impact sentence coherence and factual consistency, especially when they corrupt or remove key words or essential contextual phrases. Although the impact is less pronounced at lower levels, it escalates sharply when the frequency of deletions increases. At higher levels of deletion error, the degradation in BERTScore is similar to that seen with substitution errors, highlighting how missing characters or words disrupt structured text.
Substitution errors exhibit the most severe impact on QA performance in English and French, causing the steepest decline in BERTScore as their frequency increases. Since these errors modify characters within words, they often alter word meaning and disrupt sentence structure, making them highly detrimental to text comprehension.

However, in German, substitution errors appear to be less disruptive than in English and French. This can be attributed to the compound word structure in German, where minor substitutions can still preserve some semantic similarity. In contrast, deletions or insertions tend to fragment meaningful lexical units, making them more impactful in German than in other languages.

\textbf{Summary of findings:} Across languages, the results reveal that English and French exhibit similar degradation patterns, with BERTScore progressively decreasing as the OCR error frequency increases. However, in German, the sharpest decline is observed across all error types, particularly for substitutions and deletions. This suggests that OCR noise in German is more detrimental, probably because the older scripts have content in old German where characters such as (long s) are used instead of "s", which can often be misread as "f" or "l".

The results indicate that \textit{substitution errors are the most disruptive in English and French, while German is more affected by deletions} due to its compound word structure. \textit{Insertion errors generally cause moderate degradation}, but severe performance drops occur at high error levels. Overall, \textit{German experiences the highest performance drop}, reinforcing its greater vulnerability to OCR distortions and highlighting the need for effective OCR correction strategies.

\begin{table}[t]
\caption{Performance metrics of QA systems on English text using pre-processed and post-processed OCR input. Red values indicate the percentage drop in performance compared to using Corrected Paragraph (Ground Truth) as context.}
\label{tab:additional_study_english}
\centering
\resizebox{\columnwidth}{!}{
\begin{tabular}{llll}
\toprule
\textbf{Approach}       & \textbf{BERTScore} & \textbf{Contains}  & \textbf{EM} \\
\midrule
Corrected Paragraph (Ground Truth)     &63.65 & 57.60 & 16.20 \\
\toprule
RawOCR Paragraph        & 57.85 \textcolor{red}{(9.11\%$\downarrow$)}  &45.60 \textcolor{red}{(20.83\%$\downarrow$)}   &9.00 \ \  \textcolor{red}{(44.44\%$\downarrow$)}  \\
LLM Corrected Paragraph &59.14 \textcolor{red}{(7.09\%$\downarrow$)}  &42.80 \textcolor{red}{(25.69\%$\downarrow$)}  & 12.20 \textcolor{red}{(24.69\%$\downarrow$)}   \\
RawOCR Corrected Answer &56.10 \textcolor{red}{(11.86\%$\downarrow$)} &41.33 \textcolor{red}{(28.25\%$\downarrow$)}  & 9.16 \ \ \textcolor{red}{(43.46\%$\downarrow$)}  \\

\bottomrule
\end{tabular}
}
\end{table}

\begin{table}[t]
\caption{Performance metrics of QA systems on French text using pre-processed and post-processed OCR input. Red values indicate the percentage drop in performance compared to using Corrected Paragraph (Ground Truth) as context.}
\label{tab:additional_study_french}
\centering
\resizebox{\columnwidth}{!}{
\begin{tabular}{llll}
\toprule
\textbf{Approach}       & \textbf{BERTScore} & \textbf{Contains}  & \textbf{EM} \\
\midrule
Corrected Paragraph (Ground Truth)     &65.17& 52.40 & 14.40\\
\toprule
RawOCR Paragraph        &63.01 \textcolor{red}{(3.31\%$\downarrow$)}  &41.40 \textcolor{red}{(20.61\%$\downarrow$)}  & 12.40 \ \textcolor{red}{(13.89\%$\downarrow$)}  \\
LLM Corrected Paragraph &57.78 \textcolor{red}{(11.33\%$\downarrow$)}  &29.60 \textcolor{red}{(43.51\%$\downarrow$)}  & 5.40 \ \textcolor{red}{(62.50\%$\downarrow$)}   \\
RawOCR Corrected Answer &51.72 \textcolor{red}{(20.64\%$\downarrow$)} &19.00 \textcolor{red}{(63.74\%$\downarrow$)}  & 1.00 \ \ \textcolor{red}{(93.06\%$\downarrow$)}  \\

\bottomrule
\end{tabular}
}
\end{table}

\begin{table}[t]
\caption{Performance metrics of QA systems on German text using pre-processed and post-processed OCR input. Red values indicate the percentage drop in performance compared to using Corrected Paragraph (Ground Truth) as context.}
\label{tab:additional_study_german}
\centering
\resizebox{\columnwidth}{!}{
\begin{tabular}{llll}
\toprule
\textbf{Approach}       & \textbf{BERTScore} & \textbf{Contains}  & \textbf{EM} \\
\midrule
Corrected Paragraph (Ground Truth)     &59.80& 12.80 & 3.40\\
\toprule
RawOCR Paragraph        &55.92 \textcolor{red}{(6.49\%$\downarrow$)}  &5.60 \textcolor{red}{(56.25\%$\downarrow$)}  & 1.00 \ \ \textcolor{red}{(70.59\%$\downarrow$)}  \\
LLM Corrected Paragraph &54.27 \textcolor{red}{(9.26\%$\downarrow$)}  &3.40 \textcolor{red}{(73.44\%$\downarrow$)}  & 0.20 \ \textcolor{red}{(94.12\%$\downarrow$)}   \\
RawOCR Corrected Answer &55.21 \textcolor{red}{(7.68\%$\downarrow$)} &4.80 \textcolor{red}{(62.50\%$\downarrow$)}  & 0.60 \ \textcolor{red}{(82.35\%$\downarrow$)}  \\

\bottomrule
\end{tabular}
}
\end{table}

\vspace{-2mm}
\subsection{Effectiveness of Error Correction Strategies}
Finally, we present an additional study to mitigate the impact of OCR errors on QA through two mitigation approaches, context correction and answer correction of RawOCR text. We want to test simple solutions to reduce the effect of OCR noise in QA while preserving the syntactic and semantic integrity of the input. 
Such solutions should also be feasible to be deployed in real-world settings with limited computational and financial resources. 

We randomly sample 500 questions for each of the three languages. As the context correction approach\footnote{The prompt used for context correction. \textbf{System Prompt:} \textit{You are an expert at understanding the historical texts and correcting OCR errors.} \textbf{User Prompt:} \textit{You are provided with a historical English text containing spelling mistakes. Correct only the spelling mistakes and present the corrected text in a single paragraph. If you cannot correct the mistakes, reply with "Not able to correct."  Historical Text : \{context\}}}, we correct the RawOCR text first using the Gemma-2 27B model and then use it as context for answering questions. In contrast, in the answer correction approach, we use RawOCR text as context to generate an answer first, and then we correct it afterward. 

The context correction approach may improve performance by correcting the entire context before QA. However, it is computationally expensive due to the need to process the entire passage. On the other hand, the Answer correction approach is more lightweight and cost-efficient, as it only involves correcting short answer spans. Nevertheless, it may suffer from context misalignment, especially when the input is both noisy and lacks sufficient contextual information. We use Gemma-2 27B for both strategies as it demonstrated the strongest performance across all languages in our main experiments (Section \ref{experimental_results}). The results of this analysis are summarized for different languages in Table \ref{tab:additional_study_english}, Table \ref{tab:additional_study_french}, and Table \ref{tab:additional_study_german}, respectively. In the tables, the context correction approach is labeled as LLM Corrected Paragraph, while the answer correction approach is labeled as RawOCR Corrected Answer.

The results across English, French, and German (Tables \ref{tab:additional_study_english}, \ref{tab:additional_study_french}, and \ref{tab:additional_study_german}) consistently show that using corrected (ground-truth) paragraphs as context yields the highest QA performance across all metrics. Among the three languages, English shows the greatest benefit from pre-processing RawOCR with LLM correction, with BERTScore and EM metrics improving significantly over using RawOCR directly. In contrast, French and German show minimal or even negative gains from pre-processing, with German showing extremely low EM score with degradation of 94.12\%  compared to the ground truth. For all three languages, post-processing the answer (i.e., correcting the generated answer after QA) consistently results in the worst performance, suggesting that once the model has reasoned over noisy text, semantic distortions are difficult to recover.  Across all languages, the post-processing strategy, where the answer is corrected after being generated from RawOCR input, consistently results in the lowest performance, highlighting that once a model has reasoned over noisy context, semantic errors become difficult to correct.

These findings emphasize the \emph{importance of integrating OCR error correction early in the QA pipeline} to improve the reliability of QA systems, especially when dealing with historical texts or other archival materials of lower quality. However, given the huge collections of digitized content with vastly varying levels of OCR quality that the current memory institutions (archives, libraries, museums, etc.) held, the correction cost and effort would be enormous. It is also difficult to correct the queried texts at inference time as this would also introduce computational cost and latency in online systems. 
Therefore, more robust QA systems that are aware of OCR errors and capable of predicting correct answers based on contextual information are required.

\section{Conclusion}

In this paper, we conducted a comprehensive evaluation of QA systems under the influence of OCR-induced noise using multilingual historical texts. To facilitate this analysis, we introduced \MultiOCR, the first multilingual QA dataset derived from OCR-processed historical documents in English, French, and German. The dataset includes both RawOCR (noisy) and CorrectedOCR (clean) text, enabling controlled and comparative assessments of model robustness under real-world digitization errors.  By leveraging both CorrectedOCR and RawOCR 
we systematically analyzed how different types of OCR errors—insertions, deletions, and substitutions affect the performance of LLMs in QA. 

Our experiments reveal that OCR noise leads to substantial degradation in QA performance across all languages and metrics, with particularly severe impacts in settings with high error rates. While larger LLMs such as Gemma-2 27B and Qwen-2.5 72B show relatively better robustness, even these models experience significant accuracy drops when processing noisy text. Moreover, pre- and post-processing strategies using LLMs to correct OCR errors offer only limited benefits and can, in some cases, degrade performance even further. These findings highlight the limitations of current QA systems in handling real-world noisy text and emphasize the need for evaluation frameworks and model designs that can handle OCR distortions, especially in historical and multilingual contexts.

\textbf{Use cases of \MultiOCR:} The \MultiOCR dataset offers a unique resource to advance research on OCR-aware QA and studying QA on noisy OCR text, making it useful in several ways. It can be used to train LLMs to improve error correction capabilities and enhancing robustness against OCR inaccuracies while preserving the archaic language structure. It can also be used to expand LLMs' multilingual processing abilities by training on OCR text in multiple languages, enhancing performance in languages beyond English.


\textbf{Limitations and Future work}: While \MultiOCR includes English, French, and German, it does not encompass low-resource languages or scripts such as Latin, Finnish, and others. Future research should incorporate low-resourced languages to improve generalization and greater applicability across diverse languages. Additionally, methodologies that not only remove OCR errors but also preserve the original structure of documents could be applied. 

\begin{acks}
The computational results presented here have been achieved (in part) using the LEO HPC infrastructure of the University of Innsbruck. This Project was Co-funded by the European Union HORIZON-WIDERA-2023-TALENTS-01-01 grant 101186647 — AI4DH. Views and opinions expressed are however those of the author(s) only and do not necessarily reflect those of the European Union. Neither the European Union nor the granting authority can be held responsible for them.

\end{acks}

\section*{Usage of Generative AI}

This work involved the use of Generative AI (GenAI) models at several stages of the research process. The \MultiOCR dataset has been automatically generated using an instruction fine-tuned large language model (LLaMA 3.1-70b) for English, French, and German. The models were explicitly used for answer span extraction and question generation based on historical CorrectedOCR text. Various generative models, including LLaMA, Qwen, Mixtral, and Gemma, were used in the downstream evaluation of QA performance across corrected and noisy OCR text. Additionally, GenAI tools such as ChatGPT and Gemini were occasionally used to assist with resolving coding bugs or syntax errors in scripts or LaTeX syntax.

All conceptual work, dataset engineering, code design, and writing remain the sole intellectual contributions of the authors. GenAI tools were used in a supportive capacity, similar to that of debugging assistants or grammar checkers, without contributing novel research content.


\bibliographystyle{ACM-Reference-Format}
\balance
\bibliography{sample-base}


\end{document}